\documentclass{article}

\usepackage{PRIMEarxiv}
\usepackage{microtype}
\usepackage{graphicx}
\usepackage{subfigure}
\usepackage{booktabs} 
\usepackage{xcolor}
\usepackage{physics}
\usepackage{amsfonts}
\usepackage{amssymb}
\usepackage{bbm}
\usepackage{cite}
\usepackage[ruled]{algorithm2e}
\usepackage{bm}

\usepackage[T1]{fontenc}

\usepackage{longtable}

\usepackage{caption}
\usepackage{url}            
\usepackage{booktabs}       
\usepackage{amsfonts}       
\usepackage{nicefrac}       
\usepackage{microtype}      
\usepackage{lipsum}
\usepackage{fancyhdr}       
\usepackage{graphicx}       
\graphicspath{{media/}}     

\usepackage{orcidlink}

\usepackage{hyperref}

\hypersetup{
  colorlinks=false,
  linkbordercolor=white,
 urlbordercolor=white,
pdfborder={0 0 0}
}
\title{Risk Automatic Prediction for Social Economy Companies using Camels
\thanks{\textit{\underline{Citation}}: 
\textbf{Joseph et al., Risk Automatic Prediction for Social Economy Companies using Camels.}} 
}

\author{
  Joseph A. Gallego-Mejia \orcidlink{0000-0001-8971-4998}$^1$, Daniela Martin-Vega \orcidlink{0000-0003-3943-9849}$^2$ , Fabio A. Gonz\'{a}lez\orcidlink{0000-0001-9009-7288}$^3$ \\
  MindLab \\
  Universidad Nacional de Colombia$^{1,3}$, Universidad Distrital Francisco Jose de Caldas$^2$ \\ \\
  Bogot\'{a}, Colombia\\
  \texttt{\{jagallegom$^1$,fagonzalezo$^3$\}@unal.edu.co} dmartinv@correo.udistrital.edu.co$^2$\\ 
}

\begin{document}

\maketitle

\begin{abstract}
Governments have to supervise and inspect social economy enterprises (SEEs). However, inspecting all SEEs is not possible due to the large number of SEEs and the low number of inspectors in general. We proposed a prediction model based on a machine learning approach. The method was trained with the random forest algorithm with historical data provided by each SEE. Three consecutive periods of data were concatenated. The proposed method uses these periods as input data and predicts the risk of each SEE in the fourth period. The model achieved 76\% overall accuracy. In addition, it obtained good accuracy in predicting the high risk of a SEE. We found that the legal nature and the variation of the past-due portfolio are good predictors of the future risk of a SEE. Thus, the risk of a SEE in a future period can be predicted by a supervised machine learning method. Predicting the high risk of a SEE improves the daily work of each inspector by focusing only on high-risk SEEs. 

\keywords{Machine Learning \and Random Forest \and Forecasting \and Regression \and CAMELS \and Risk Prediction.}
\end{abstract}
\section{Introduction}
Reviewing, monitoring and inspecting are important tasks for governments when evaluating social economy enterprises (SEEs). Social economy enterprises are a group of private or semi-private enterprises in which profit distribution and decision making are not related to the asset ownership of each member \cite{julia2003social}. Governments have to inspect these SEEs and are responsible for categorizing and measuring companies that have medium-high risk and medium-high severity. Risk relates to the likelihood of a company going into financial failure, i.e., being unable to pay its debts. Severity measures the relative impact of a potential SEE bankruptcy within a relative area, e.g., a town, city or state. It should be noted that all SEEs cannot be inspected on-site due to limitations on the number of inspectors in a given time. Therefore, a good analysis finding medium-high risk SEEs should be performed by supervisors without an on-site inspection. In this paper, we present a model that measures the future risk of financial failure of an SEE using: its financial history; operational reports; the CAMELS rating system, which is a system created in the U.S. to evaluate the overall condition of its banks; and macroeconomic indicators, such as the consumer price index.

This paper presents a novel model to predict the risk of financial failure of an SEE in the next period considering historical financial and operational data. This predictive model is built using machine learning models with the help of historical financial and operational data provided by each SEE. A preprocessing step was performed in order to clean and transform the raw data. This new prototype model shows the opportunities offered by machine learning for the decision making process. Our main hypothesis is that the new model will be a valuable tool for prioritizing which SEE requires inspection by government inspectors.\\

The rest of this paper is organized as follows. Section 2 reviews related work on bankruptcy prediction and the CAMELS system. Section 3 presents the description of the data. Section 4 presents the description of the proposed model. Section 5 presents the experimental evaluation of the proposed model. Section 6 presents the results and discussions achieved with the proposed model. Finally, Section 7 presents the conclusions.

\section{Background and Related Work}
Social economy enterprises are really important for healthy societies; they provide several jobs and sustain part of the economy. Assessing their performance and possible bankruptcy is a crucial task for governments. Therefore, several machine learning methods have been applied to predict bankruptcy \cite{Hernandez2013,Olson2012,Kim2010}. These algorithms are based on financial figures such as balance sheet and income statements, company-specific variables and stock market data. In addition, for a more general evaluation of a company, it is necessary to evaluate the qualitative attributes of the company \cite{Chmielewska2018}. \\

Several types of methods for predicting bankruptcy have been proposed in the state-of-the-art. In \cite{Gepp2010}, the authors used decision trees to predict whether a given firm will go bankrupt in the following year. They showed that decision trees are better than logit regression.  In \cite{Khermkhan2015}, the authors built a model to predict the distress of a given company. They used logit, probit, multivariate discriminant analysis and artificial neural networks, showing that logit and probit are good, as they have explainable and understandable properties. In contrast, artificial neural networks performed well but had poor explainable and understandable properties. In \cite{Dima2016}, the authors evaluated 3,000 companies in Romania in eight categories, from AAA to D. They tried to predict their probability of downward transition from one year to the next. They used logit regression and artificial neural network and showed that artificial neural networks perform better than logit regression. In \cite{Tobback2017}, the authors supplemented traditional data, such as financial figures, with quality data, such as a company sharing director or senior managers. With these relationships, a neighborhood prediction model was built. They showed that when quality data is combined with financial data, the performance of the algorithm increases. In \cite{Chmielewska2018}, the authors evaluated four machine learning techniques: decision trees, neural networks, random forest, and logistic regression. They showed that the random forest outperforms the other methods for bankruptcy prediction.    \\

Financial datasets pose various problems, for example, the ratio of bankrupt to non-bankrupt companies is imbalanced, i.e., there are more non-bankrupt companies. In this case, the algorithm cannot be evaluated with the accuracy metric alone and more robust metrics such as f1-score, precision or recall can be used. Therefore, in \cite{zoricak2020}, the authors propose a comparison between three algorithms that deal with imbalance: probabilistic least-squares classification for outlier detection, an isolation forest, and a one-class support vector machine. They show that probabilistic least-squares classification (LSDA) outperforms the other two methods. The size of the company is also a really important factor to take into account. Another problem is to evaluate small, medium and large companies with the same model. In \cite{Elkalak2016}, the authors showed that firm size is a good predictor variable for the success or failure of a firm \cite{Gepp2010}.

The CAMELS model was developed in the United States in 1991 to assess the overall performance of banks through capital adequacy, assets, management capacity, earnings, liquidity and sensitivity to market risk. This model gives the overall health and performance on a rating system between one and five, where one implies safe performance and five implies unsatisfactory performance, of the banking system, but can be used to evaluate other types of companies such as SEEs \cite{fdic}. In \cite{Gilbert2002CouldAC}, the authors developed an advanced CAMELs model with a Supervisory Risk Assessment and Early Warning Systems to evaluate banks over a two-year period. In \cite{Whalen2005AHM}, the authors developed a new method to calculate the probability of migrating from a low to a high level of risk based on CAMELS. In \cite{derviz2008predicting}, the authors evaluate the possibility of predicting changes in bank ratings in the following years by showing that they could predict some variables such as capital adequacy, and leverage. In \cite{dincer2011performance}, the authors used the CAMELS system to evaluate the performance of Turkish banks in the period from 2001 to 2008, showing that a strong liquidity ratio signifies overall good health. In \cite{nandi2013comparative}, the authors analyzed the development of the public and private bank sector in India. They showed that CAMELS is a good rating system for assessing bank performance. In \cite{kaur2015financial}, the authors evaluate several Indian banks using CAMELS and showing that 95\% of the change is given by debt-to-equity ratio, loan-to-deposit ratio, income per employee, capital adequacy ratio, and total investment-to-total assets ratio.

\section{Data Description}

The Colombian government collects data from each SEE at regular intervals. Some companies have to send information monthly, others every six months. The type of data collected are total assets, total savers, total employees, total associates, financial portfolio per debtor, total income in the period, among others. Each company receives a risk label ranging from one to five, where one indicates low risk and five indicates high risk. The process of labeling each company is performed manually by an inspector. Note that the risk assesses the probability that a new company will default in the next period. In this work, the risk is the value we want to learn to predict with an automated algorithm. \\

To build the predictive model, we used historical data collected from each SEE provided by the government. The periodicity of the financial reports that each SEE has to submit to the Colombian government varies according to the number of employees, capital, assets, among others. However, all SEEs have to submit their financial reports every semester. For this reason, we selected a semi-annual periodicity for our proposed model. We use semi-annual periodicities from the first half of 2016 to the first half of 2019.

\begin{table}[hbt!]
\caption{Number of social economy enterprises (SEEs) that submitted financial information for four consecutive periods.}
\centering
\begin{tabular}{|c|c|}
\hline
Consecutive  & Numbers of SEEs that presented\\
Periods &  financial information \tabularnewline
\hline
\hline
2016-1 al 2017-2 & 3287\tabularnewline
\hline
2016-2 al 2018-1 & 2056\tabularnewline
\hline
2017-1 al 2018-2 & 1999\tabularnewline
\hline
2018-1 al 2019-1 & 2042\tabularnewline
\hline
\end{tabular}
\end{table}

The data is very unbalanced due to intrinsic properties. The majority of companies belong to risk 2 or 3, with 46.6\% and 51.8\% respectively. There are a similar number of companies belonging to risks 1 and 4 with 0.2\% and 0.1\% respectively. Finally, risk 5 has the lowest number of companies with 0.03\%. We used an oversampling technique for companies with risk 1, 4 and 5 and applied an undersampling to companies with risk 2 and 3. The method used to balance the data was SMOTE, which has good results in the literature \cite{chawla2002smote}. 

\section{Model Description}

We proposed a supervised machine learning model. A supervised machine learning model takes an input data $X$ and learns to predict an output $\hat{y}$ that is compared to the true value of $y$. In the present model, the input data is three consecutive periods of historical financial and operational data given by each SEE and the output prediction is the fourth consecutive period of the same historical data. \\

\begin{equation}
\label{eq:autorregressive_model}
X_t = c + \sum_{i=1}^p X_{t-i} + \epsilon_t
\end{equation}

We transformed the data as follows: first, historical data from three consecutive periods of the same SEE were concatenated into a single row; second, historical data from the fourth consecutive period are selected as the period to be predicted; finally, we proposed a supervised machine learning model where the model learns a function that predicts the fourth consecutive period given the three consecutive periods. The equation \ref{eq:autorregressive_model} presents a mathematical notation of the model representation. Each company was characterized by a set of characteristics that change over time denoted $X_t^j$ where t refers to time and j refers to the characteristic. For each SEE $i$, we have $j$ features for consecutive periods $t-3$, $t-2$ and $t-1$. A supervised machine learning model predicts the risk for period $t$.  \\

Furthermore, we hypothesize that the use of macroeconomic variables for each period can increase the power of the model. Therefore, we use the following macroeconomic variables: consumer price index, unemployment rate and gross domestic product.

\section{ \label{sec:experimenta-evaluation} Experimental Evaluation}

We developed three slightly different models and compared them to an assessment metric. Our goal was to have a prediction model that could predict risk at each of the risk levels 1 to 5 according to the CAMELS risk model. For each model, we evaluated several regression models: random forest, support vector machines, logistic regression and neural networks.

\subsection{Experimental Setup}

\begin{table}
\centering
\caption{\label{tab:dataset} Data description: total number of data, number of training points and number of test points }
\begin{tabular}{|c|c|c|}
\hline
Total & Training (70\%) & Test (30\%)\tabularnewline
\hline
\hline
9383 & 6568 & 2815 \tabularnewline
\hline
\end{tabular}
\end{table}

Cross-validation with stratification according to SEE risk was used to select the best model. A 70-30 partition was used for cross-validation. Seventy percent of the data were used as training points and thirty percent were used as test points. The table \ref{tab:dataset} shows the number of training and test data points. Each categorical variable was transformed as a dummy variable or one-hot encoding \cite{hardy1993regression}. Eighty variables were used to build the model, such as total liabilities for the period, total deposits, total equity, among others. Table \ref{table:variables} shows the complete list of variables used. Three types of preprocessing tools were used for the ordinal variables: standardization, or mean removal and variance scaling; min-max scaling; and log-normal. The best result was obtained with log-normal preprocessing. We used four different supervised methods for this problem. The methods we used were: random forest, support vector machines (SVM), logistic regression and neural networks. The results presented in the following section are based on random forest. Random forest was selected because of its good performance compared to the other methods in terms of accuracy. Random forest is a supervised method that ensembles several trees. It uses a voting system to select which decision trees to use for the given problem \cite{Pavlov2019}. A replication method was used to select the trees that the Random Forest considered. Support vector machines was trained using a Gaussian kernel. A random grid was used for hyper parameters tuning of each algorithm. The parameters that were optimized by the search were: number of neurons, number of hidden layers, number of iterations, $\gamma$-parameter, $C$ complexity parameter, number of decision trees, maximum depth of each decision tree, number of openings in each node of the decision trees and the minimum number of openings in each leaf of the decision trees.

\begin{figure*}[tbh]
\begin{centering}
\includegraphics[scale=0.43]{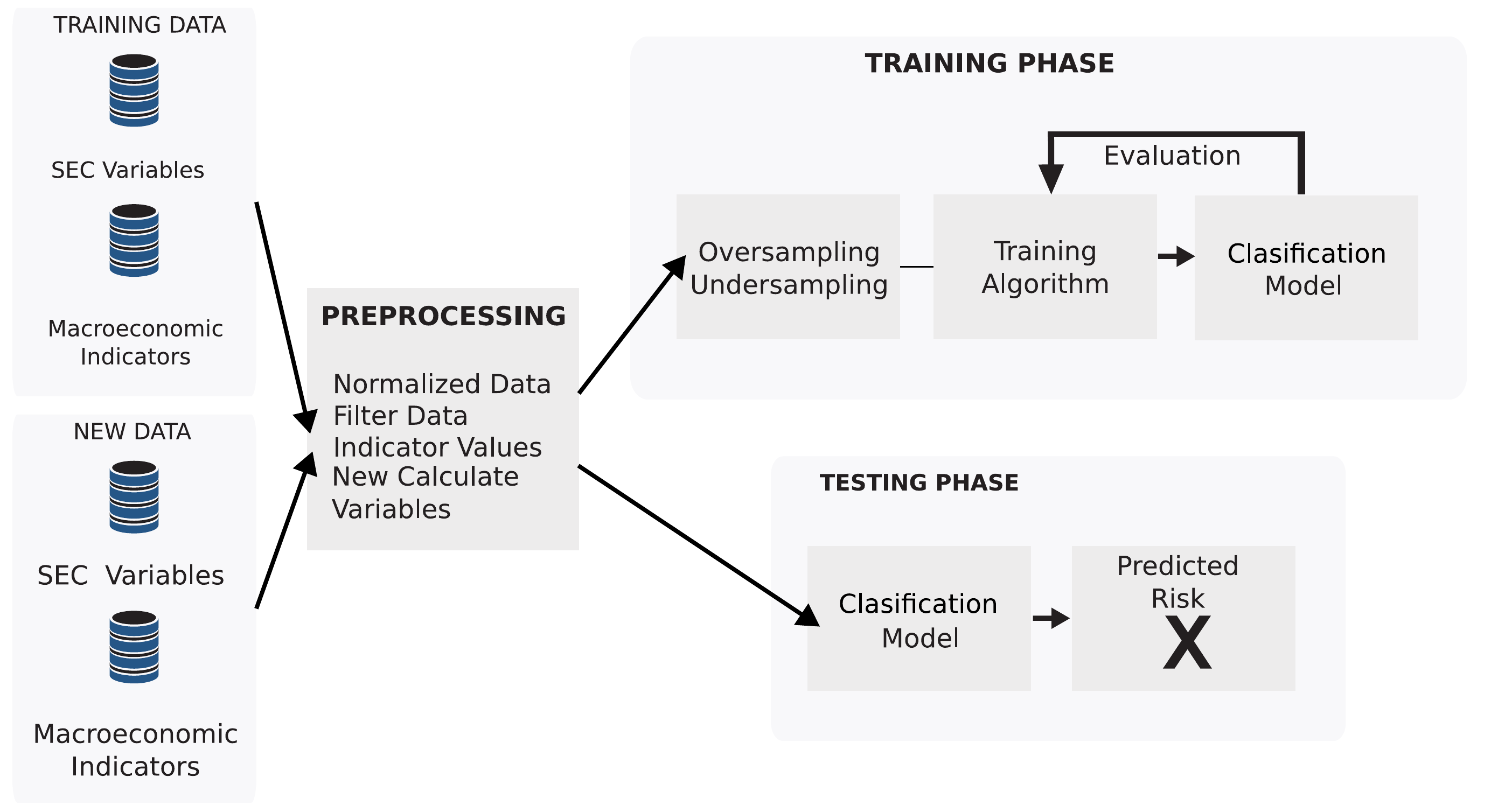}
\par\end{centering}
\caption{\label{fig:diseno-experimental}Experimental Setup}
\end{figure*}

Three models were evaluated. The first model takes into account SEE variables, such as chart of accounts, number of debtors, number of employees, and legal nature, among others. In addition, this first model is naive and did not use feature extraction, which is commonly used in machine learning models. The second model used the variables of the first model and additionally the variation between two consecutive periods of several variables such as number of associates, number of employees, total assets, among others. The third model shows the addition of the CAMELS variables and macroeconomic indicators (MI).

As an evaluation measure, we used the precision of each risk class in a confusion matrix. We created a confusion matrix for each of the experiments. Each row of the confusion matrix corresponds to the actual risk value and each column corresponds to the risk predicted by the algorithm. The accuracy is calculated as the ratio between the companies with correct predicted risk $i$ versus all companies belonging to risk $i$. The matrix allows visualizing the values at which the prediction algorithms make an error, i.e. the prediction model predicts that the risk is $i$ but the actual risk is $j$. The evaluation equation is $\text{Accuracy}_{\text{risk}_{i}}$ $=$ Number of companies that was predicted as risk $i$ and the real risk is $i$ $/$ Number of companies predicted in risk $i$. The predicted model is expected to have the lowest possible error when comparing the predicted risk with the actual risk. For this experiment, it is important to consider prediction models that maximize accuracy when the risk is higher. Since high-risk SEEs are the main concern of the government. Those high-risk SEEs with medium-high severity may affect a large number of people and, consequently, may affect the country's economy. For the latter reason, the model needs to have high confidence in those SEEs that have high risk.

\section{Results and Discussions}
\begin{table}
\centering
\caption{\label{tab:precision-result}Precision results for each proposed model}
\begin{tabular}{|c|c|c|c|c|}
\hline
Model & Random  & Neural  & SVM & Logist \\
&Forest & Networks & & Regression
\tabularnewline
\hline
\hline
Chart of account (CA) +  & $\bm{72\%}$ & 69.3\% & 70.1\% & 65.5\% \\
Descriptive variables (DV) & & & & 
\tabularnewline
\hline
CA + DV +  & $\bm{70.95\%}$ & 68.3\% & 68.2\% & 63.5\%\\
Variation between periods (VBP)  & & & & 
\tabularnewline
\hline
CA + DV + VBP + & $\bm{76.66\%}$ & 75.5\% & 74.5\% & 70.3\%\\
CAMELS/RISK + MI   & & & & 
\tabularnewline
\hline
\end{tabular}
\end{table}

Table \ref{tab:precision-result} shows the precision result for each predicted model using different classification models. The best classification method in each model is random forest. Therefore, the following analysis uses only the random forest as a classification method. The first model achieved an overall accuracy of 72\%. Table \ref{fig:confusion-matrix-1} shows the confusion matrix for the first proposed model. The fourth highest risk achieves only 9\% accuracy. Most of the SEEs fall between risk two and three. However, the assumption for the following proposed models is that the accuracy for the fourth risk can be improved.

\begin{table}[]
\centering
\caption{Confusion matrix of the first model}
\label{fig:confusion-matrix-1}
\begin{tabular}{|c|c|c|c|c|c|}
\hline
           & \textbf{1}    & \textbf{2}    & \textbf{3}    & \textbf{4}   & \textbf{5}   \\ \hline
\textbf{1} & \textbf{14\%} & 79\%          & 7\%           &              &              \\ \hline
\textbf{2} &               & \textbf{80\%} & 20\%          &              &              \\ \hline
\textbf{3} &               & 33\%          & \textbf{67\%} & 1\%          &              \\ \hline
\textbf{4} &               & 7\%           & 84\%          & \textbf{9\%} &              \\ \hline
\textbf{5} &               & 100\%         &               &              & \textbf{0\%} \\ \hline
\end{tabular}
\end{table}

The second proposed model did not improve accuracy. This model achieved only 70.95\% accuracy. Table \ref{fig:confusion-matrix-3} shows the confusion matrix of the third proposed model. This third proposed model achieves an overall accuracy of 76.66\%. In this case, our model can predict with 18\% accuracy the fourth risk. In this model, we found that if the legal nature of the SEE is an employee fund or a cooperative, the risk of SEE increases. In addition, we found that the variation of past-due portfolio between consecutive periods is a good predictor of SEE risk. We selected the third model as the best because of its overall accuracy and good results in predicting the risk of the fourth period.

\begin{table}[]
\centering
\caption{Confusion matrix of the third model}
\label{fig:confusion-matrix-3}
\begin{tabular}{|c|c|c|c|c|c|}
\hline
           & \textbf{1}    & \textbf{2}    & \textbf{3}    & \textbf{4}    & \textbf{5}   \\ \hline
\textbf{1} & \textbf{25\%} & 75\%          &               &               &              \\ \hline
\textbf{2} &               & \textbf{78\%} & 22\%          &               &              \\ \hline
\textbf{3} &               & 22\%          & \textbf{77\%} &               &              \\ \hline
\textbf{4} &               &               & 82\%          & \textbf{18\%} &              \\ \hline
\textbf{5} &               & 100\%         &               &               & \textbf{0\%} \\ \hline
\end{tabular}
\end{table}

The proposed prediction model for the government showed good results in predicting the fourth risk of an SEE. This model could be improved if we add new variables and/or feature extractions. In addition, we could add more variables from the chart of accounts. In the proposed prediction model we did not take into account whether or not the government inspected the SEE. This information could help predict the real risk of some SEEs. \\

The current forecasting model could be improved if we took a larger data window than the last 10 years. A larger window has the risk of adding noise to the model, as the model could learn patterns that are not necessarily true with the current market dynamics. However, there are a large number of cyclical patterns within the economy, so using a larger window could be a good opportunity to improve the accuracy of the model.

\section{Conclusions}
Governments have to supervise and inspect social economy enterprises. These enterprises present different levels of risk depending on their management and finances. In this paper, we seek to extract knowledge from historical data provided by each social economy enterprise. The proposed best prediction model presented in section \ref{sec:experimenta-evaluation} aims to show the impact that machine learning models can have on the management decision process. We show that we can build a prediction model using data stored on government servers. This model shows that using this data we can predict whether a social economy enterprise will enter a higher risk in the next period. The best machine learning method for this model is random forest. In addition, the model is very good at predicting the fourth risk. With this model, the government could improve the management of its inspectors. For future work, more accounting plan variables could be included. In addition, it could be included whether an SEE was previously inspected by the government.

\appendix
\section{Variables Used}


\label{table:variables}
\begin{itemize}
\item Type of organization   
\item Whether it is a cooperative, fund or other type of organization.    
\item Type of company: Multi-active, employee funds.                      
\item Type of supervision: 1,2,3.                                         
\item Group Niif of the organization                                      
\item Department of the organization                                      
\item Municipality of the organization                                    
\item Category of the organization                                        
\item Number of associates                                                
\item Number of employees                                                 
\item Number of offices                                                   
\item Number of correspondents                                            
\item Number of savers                                                    
\item Number of debtors                                                   
\item Total assets of the organization                                    
\item Client portfolio for the period                                     
\item Net client portfolio for the period                                 
\item Clients' consumer portfolio for the period                          
\item Clients' housing portfolio for the period                           
\item Customer commercial portfolio for the period                        
\item Clients' micro portfolio for the period                             
\item Total investments of the organization                               
\item Cash receivable under agreement                                     
\item Total liabilities for the period                                    
\item Total deposits                                                      
\item Total deposits in bank accounts                                     
\item Total CDT deposits                                                  
\item Total contractual deposits                                          
\item Total permanent savings deposits     
\item Total equity                                                        
\item Total social contributions                                          
\item Total surplus                                                       
\item Total income for the period                                         
\item Total expenses for the period                                       
\item Total gross portfolio                                               
\item Total past due portfolio                                            
\item Total female members                                                
\item Total male members       
\item Consolidated risk rating                                            
\item Risk rating                                                         
\item Camel rating                                                        
\item Credit risk                                                         
\item Liquidity risk                                                      
\item Operational Risk                                                    
\item Sarassoft risk                                                      
\item Total capital of the organization                                   
\item Total assets of the organization  
\item Total administration expenses                                       
\item Total profitability of the organization                             
\item Organization's liquidity                      
\item Variation between two consecutive periods of associates             
\end{itemize}
\vfill\null
\begin{itemize}
\item Variation between two consecutive periods of employees              
\item Change between two consecutive periods in the number of offices     
\item Change in the number of savers between two consecutive periods      
\item Variation between two consecutive periods in the number of debtors. 
\item Change between two consecutive periods in total assets           
\item Variation between two consecutive periods of gross portfolio  
\item Variation between two consecutive periods of the consumer portfolio 
\item Variation between two consecutive periods of the housing portfolio  
\item Change between two consecutive periods for the commercial portfolio 
\item Change in micro portfolio between two consecutive periods           
\item Change between two consecutive periods in the investment portfolio  
\item Variation between two consecutive periods of covenants receivable   
\item Variation between two consecutive periods of liabilities            
\item Change in deposits between two consecutive periods                  
\item Change between two consecutive periods in demand deposits           
\item Variation between two consecutive periods of CDT deposits           
\item Variation between two consecutive periods of contractual deposits   
\item Variation between two consecutive periods of permanent savings deposits                                                     
\item Variation between two consecutive periods of shareholders' equity   
\item Variation between two consecutive periods of social contributions   
\item Variation between two consecutive periods of the surpluses          
\item Variation of income between two consecutive periods                 
\item Change in expenses between two consecutive periods                  
\item Variation between two consecutive periods of overdue accounts receivable                                       
\item Variation between two consecutive periods of female associates      
\item Variation between two consecutive periods of male associates        
\item Variation between two consecutive periods for other associates      
\item Risk weighting (variable to be predicted)
\end{itemize}



\section{Acknowledgements}
This work was partially funded by ``Contrato interadministrativo 232" between \textit{Universidad Nacional de Colombia} and \textit{Supersolidaria}.

\bibliography{paper}
\bibliographystyle{splncs04}

\end{document}